\RequirePackage{fix-cm}
\documentclass{rob}
\usepackage{graphicx}
\usepackage[hidelinks,bookmarks=false]{hyperref} 
\usepackage[caption=false]{subfig}
\usepackage{adjustbox}
\usepackage{cite}
\usepackage{amsmath,amssymb,amsfonts}
\usepackage{algorithmic}
\usepackage{textcomp}
\usepackage{xcolor}
\usepackage{graphics} 
\usepackage{epsfig} 
\usepackage{color}
\usepackage{booktabs} 
\usepackage[binary-units=true]{siunitx} 

\DeclareSIUnit\px{px}

\def\BibTeX{{\rm B\kern-.05em{\sc i\kern-.025em b}\kern-.08em
    T\kern-.1667em\lower.7ex\hbox{E}\kern-.125emX}}

\doi{03.2020/xxxx}
\volume{}
\issue{}
\pubyear{2020}
\cpyear{2020}
\endpage{12}

\begin{document}

\title[WGANVO: Monocular VO based on GAN]{WGANVO: Monocular Visual Odometry based on Generative Adversarial Networks}


\author{Javier Cremona$\dagger$\thanks{Corresponding author. E-mail:
cremona@cifasis-conicet.gov.ar}, Lucas Uzal$\dagger$ and Taih\'u Pire$\dagger$}
\affil{$\dagger$CIFASIS (CONICET-UNR)}

\ADaccepted{MONTH DAY, YEAR. First published online: MONTH DAY, YEAR}


\maketitle

\begin{summary}
In this work we present WGANVO, a Deep Learning based monocular Visual Odometry method. In particular, a neural network is trained to regress a pose estimate from an image pair. The training is performed using a semi-supervised approach. Unlike geometry based monocular methods, the proposed method can recover the absolute scale of the scene without neither prior knowledge nor extra information. The evaluation of the system is carried out on the well-known KITTI dataset where it is shown to work in real time and the accuracy obtained is encouraging to continue the development of Deep Learning based methods.

\end{summary}

\begin{keywords}
Visual Odometry; Localization; Deep Learning; GAN.
\end{keywords}

\section{Introduction}
\label{sec:introduction}
Giving an autonomous robot the ability to accurately estimate its pose is a fundamental task in the field of mobile robotics \cite{thrun2005probabilistics,siciliano2016handbook}. This problem is known in mobile robotics as \emph{Localization}. At the same time, \emph{Odometry} is defined as the process of estimating the motion of a robot through time, based on measurements obtained from its sensors.

Several sensors, such as inertial measurement units (IMU), odometers, lasers and cameras, have been used to calculate the odometry. Cameras are sensors of great interest given their low cost, portability and the rich information they provide from the observed scene. When cameras are used to calculate odometry, the technique is called Visual Odometry (VO) \cite{nister2004visual,scaramuzza2011visual}.

VO has become an object of study in recent times, achieving great results due to the advances in the field of Computer Vision \cite{nister2004visual,comport2010realtime, krombach2016combining, engel2018dso, forster2014svo}. In general, traditional VO methods are based on geometry to estimate the robot pose \cite{pire2017sptam}.

On the other hand, in the last decade, Convolutional Neural Networks (CNN) have been successfully applied to solve several Computer Vision problems \cite{lecun2015deep}, for instance, object detection, object classification, semantic segmentation, and feature extraction \cite{salimans2016improved,krizhevsky12alexnet}. In particular, Generative Adverse Networks (GAN) \cite{goodfellow2014generative} are of great interest, since it has been shown that they are a successful technique for training deep CNNs in an unsupervised manner. Through this methodology, two networks are trained simultaneously. A network, called Generator, is capable of generating synthetic samples similar to the ones from the dataset, while, at the same time, another network, called Discriminator, discriminates real samples from those generated by the Generator. The Discriminator captures high level representations of data abstraction without labeled data and can be reused in a semi-supervised scheme to solve other tasks, such as multi-class classification \cite{salimans2016improved, radford2015unsupervised}.
\begin{figure}[!tbp]
  \centering
  \includegraphics[width=\columnwidth]{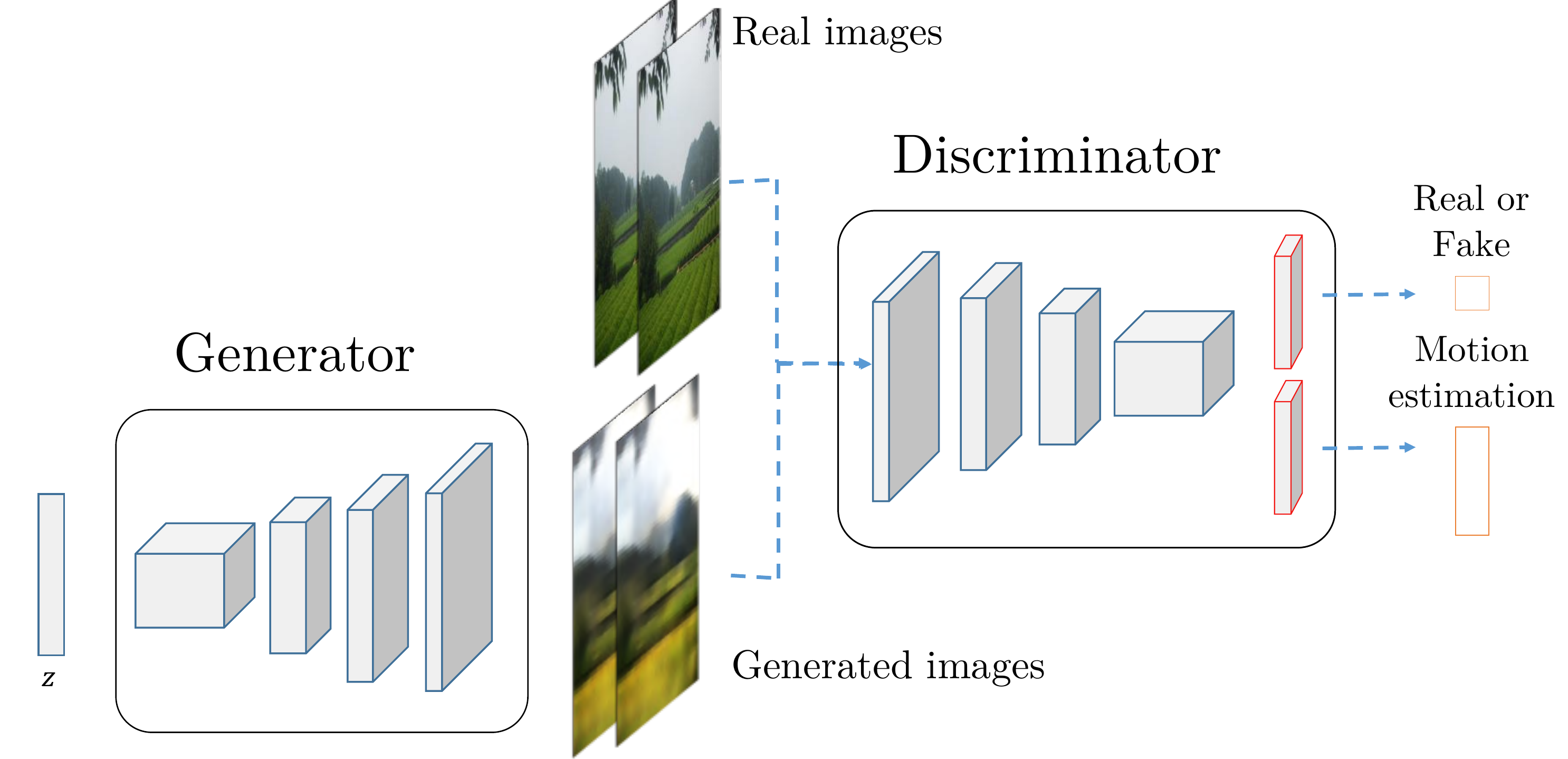}
  \caption{Architecture of the proposed method.}
  \label{fig:our-wgan}
\end{figure}

In recent years, CNNs have been applied to solve the VO problem implementing an \emph{end-to-end} approach \cite{wang2017deepvo,tateno2017cnn,agrawal2015learning}. This approach proposes using a convolutional neural network to solve a task completely, rather than dividing it into different stages. Such a division is common in traditional VO methods \cite{murartal2015orb, engel2018dso}, which are usually composed of different stages such as camera calibration, feature detection, feature matching, motion estimation, and local optimization.

In this work we introduce WGANVO, a GAN-based CNN that is trained to solve the VO problem in an end-to-end manner. In particular, our network is based on a variant of GAN known as WGAN-GP \cite{gulrajani2017improved}. The proposed Discriminator regresses the relative motion between an image pair it receives as input. The semi-supervised approach used for training combines the unsupervised technique of GAN with labeled data. The adversarial training makes the Discriminator capture as much information as possible about the images beyond the supervised signal backpropagated from the prediction error. The proposed architecture can be seen in Figure~\ref{fig:our-wgan}. Evaluations on the public dataset KITTI \cite{geiger2013vision} show that the proposed method works in real time and with promising results. The source code will be released for the benefit of the Robotics and Deep Learning community\footnote{\url{https://github.com/CIFASIS/wganvo}}.

The remainder of this paper is organized as follows: Section \ref{sec:related} reviews related work. We describe our method in Section \ref{sec:method}. Furthermore, the semi-supervised approach is introduced together with a loss function for the supervised part. Experimental results of the proposed method are given in Section \ref{sec:experiments}. These results include images generated by the network, a comparison with a state-of-the-art method and different types of training. We close the paper with conclusions and an outlook on future work in Section \ref{sec:conclusions}.

\section{Related Work}
\label{sec:related}
Most of state-of-the-art algorithms are geometry based methods. They are usually divided into feature based methods \cite{murartal2015orb, murartal2017orb, pire2017sptam} and direct methods \cite{engel2018dso, forster2014svo, engel2014lsd}. In recent years, Deep Learning techniques have been applied in different ways in order to solve the VO problem. On the one hand, methods such as DVSO (Deep Virtual Stereo Odometry) \cite{yang2018deep} use a CNN to simply solve some module of the pipeline. In particular, DVSO is strongly based on DSO \cite{engel2018dso}, a \emph{direct} method that operates on pixel intensities by minimizing photometric error along with a joint optimization of all model parameters. DVSO implements a CNN to predict depth maps when starting a new keyframe. In addition, this prediction adds geometric constraints to the joint optimization.

Another approach used is to fully address the VO problem through Deep Learning techniques. This is the approach chosen in this paper, and it is known as \emph{end-to-end}. The main advantage is that, unlike traditional methods, it requires no effort to invest in each of the modules to ensure good performance. On the other hand, traditional monocular methods are not able to estimate the absolute scale of the world without using some prior knowledge or extra information. Monocular methods that apply Deep Learning techniques are not affected by this problem since a CNN can learn what size objects are in real world during the training stage.

DeepVO \cite{wang2017deepvo} addresses the problem in an end-to-end manner. Its authors propose to train a deep recurrent convolutional neural network. The convolutional part of the network captures different characteristics and patterns of the different images. At the same time, the recurrent part models the motion of the camera from an image sequence. Recurrent networks achieve a great performance when they capture information from a sequence since they have information from previous motion at the moment of analyzing an image pair. Unlike our work, DeepVO uses a supervised training scheme, which requires a huge amount of images and labeled data in order to correctly generalize over unseen data.

UnDeepVO \cite{li2017undeepvo} is another example of an end-to-end method. The authors propose an architecture based on two CNNs: one CNN receives two consecutive monocular images and estimates the camera motion, while the other network generates depth maps from the input. The training uses a novelty cost function that allows to work in an unsupervised manner. During the training stage, stereo images are needed to recover the absolute scale of the scene and generate depth maps, even though it is a monocular method. 

For its part, GANVO \cite{almalioglu2018ganvo} is an unsupervised VO method that combines generative adversarial networks with DeepVO and UnDeepVO. The Generator builds depth maps, while a recurrent convolutional network takes an image pair as input and estimates the pose. Given the depth information and the estimated pose, a synthetic image is generated from one of the images of the incoming pair. The goal is to generate an image as similar as possible to the other image of the pair. Finally, the Discriminator receives pairs of frames, both real images and synthetically generated pairs. The Discriminator output is a probability value indicating whether the pair is real or synthetic. From this output, the Generator and the recurrent network learn to be more precise in their task. GANVO requires more information, such as depth maps, and presents a more complex architecture that includes recurrent convolutional networks.

A more complete and complex approach to the problem is that presented by GeoNet \cite{yin2018geonet}. The authors divide the problem into the reconstruction of the static scene on the one hand, and the location of dynamic objects on the other. To do this, they combine information from depth, pose and optical flow, implementing a CNN to solve each of these sub-tasks. Remarkably, GeoNet shows great precision in each of them.

PoseNet \cite{kendall2015posenet} addresses the \emph{Global Localization} problem. In this case, the location of the camera in a known scene must be estimated from an image. PoseNet implements a CNN, whose architecture is based on GoogLeNet \cite{szegedy2015going} and uses more than 20 convolutional layers. This method works in real time, showing good performance in both indoor and outdoor environments and in unfavourable weather conditions.

\section{Proposed Method}
\label{sec:method}
This section introduces WGANVO, the proposed method for solving the VO problem using CNNs. Our method uses consecutive pairs of frames of a monocular camera as input and returns the estimated motion between them.

\begin{figure*}[!tbp]
    \centering
    \includegraphics[width=\textwidth]{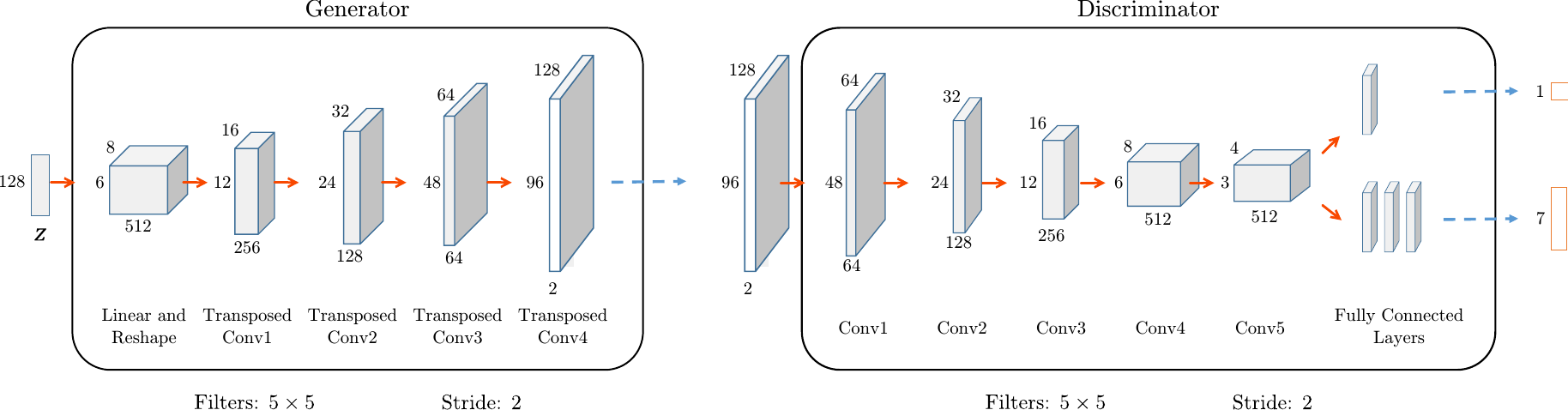}
\caption{WGANVO architecture based on WGAN-GP \cite{gulrajani2017improved}. The Discriminator input is a pair of frames and its output is adapted to estimate the 7 components that represent the motion between the input frames (a quaternion and a translation vector).}
    \label{fig:wganvo-conf}
\end{figure*}

The resulting model is based on a generative adversarial network known as WGAN-GP \cite{gulrajani2017improved}. The architecture used is shown in Figure~\ref{fig:wganvo-conf}. The Generator receives a random vector ($\mathbf{z}$) as input sampled from a normal distribution. Given this input, convolutional layers are applied, and finally, two synthetic images are returned. The Discriminator receives a pair of consecutive frames of a sequence as input. Subsequently, convolutional layers are used to extract features from the incoming images. The Discriminator is then split into two blocks. On the one hand, there is a linear layer, whose output is the probability that the input is a pair obtained from the training dataset (in contrast to a pair generated by the Generator). On the other hand, there are three fully-connected layers, which estimate the relative transformation between the two input frames. Instead of representing this transformation using a SE(3) matrix, we choose to use a translational vector and a unit quaternion for the rotational part \cite{kendall2017geometric}. Quaternions are a good alternative in optimization problems, since they are parameterized with 4 values against the 9 values that a rotation matrix of SO(3) requires. In addition, it is simpler to ensure that the quaternion has unit norm than the orthogonality of a rotation matrix.
\subsection{Training}
As mentioned above, the Generator takes a random vector of 128 components as input, sampled from a uniform distribution and its output is a pair of consecutive frames. The Discriminator is initially fed with images pairs generated by the Generator and then with images pairs from a real dataset. On the one hand, it is expected that the Discriminator learns high level details by training him to separate real frames from those that are generated. On the other hand, it is trained to solve the VO problem.

Labeled training samples are pairs of the form of $\left(\left(I_{i},I_{i+1}\right),\left(\mathbf{x}_{i(i+1)},\mathbf{q}_{i(i+1)}\right)\right)$, where $I_{i}, I_{i+1}$ are consecutive frames of a video sequence, $\mathbf{x}_{i(i+1)}$ is a translation vector in $\mathbb{R}^{3}$ between the times when $I_{i}$ and $I_{i+1}$ were captured, and $\mathbf{q}_{i(i+1)}$ is the corresponding orientation change, represented by a unit quaternion. Because ground-truth poses are stored as elements of $\mathrm{SE}\left(3\right)$, during data loading the rotational part of each relative transformation $\mathbf{T}_{i(i+1)}$ is taken and converted to the corresponding unit quaternion $\mathbf{q}_{i(i+1)}$, while the translational part of $\mathbf{T}_{i(i+1)}$ results in $\mathbf{x}_{i(i+1)}$. Finally, the output of the adapted Discriminator is composed of the estimate of the translational vector $\hat{\mathbf{x}}_{i(i+1)}$, and the estimate of the corresponding quaternion $\mathbf{\hat{q}}_{i(i+1)}$ (in addition to the probabilistic output of the Discriminator).

\subsection{Loss function}
 
The supervised part of the training implements a cost function that minimizes the rotational part and the translational part separately. The translational part is defined in equation \ref{eq:tr_part}:
\begin{equation}\label{eq:tr_part}
L_{\mathbf{x}}\left(I_{m},I_{n}\right)=\left\Vert \mathbf{x}-\mathbf{\hat{x}}\right\Vert _{2},
\end{equation}
where $\mathbf{x}$ is the position change associated with $\left(I_{m},I_{n}\right)$ and $\mathbf{\hat{x}}$ is the change of position estimated by the network when the input is the pair of frames $\left(I_{m},I_{n}\right)$. On the other hand, the rotational part is defined in equation \ref{eq:rot_part}:
\begin{equation} \label{eq:rot_part}
L_{\text{\ensuremath{\mathbf{q}}}}\left(I_{m},I_{n}\right)=\left\Vert \mathbf{q}-\mathbf{\hat{q}}\right\Vert _{2},
\end{equation}
where $\mathbf{q}$ is the change of the orientation associated with $\left(I_{m},I_{n}\right)$ and $\mathbf{\hat{q}}$ is the change of the orientation estimated by the network when the input is $\left(I_{m},I_{n}\right)$. Finally, the following cost function is defined in equation \ref{eq:cost_f}, based on the work of Kendall et al. \cite{kendall2017geometric}:
\begin{equation} \label{eq:cost_f}
L_{\beta}\left(I_{m},I_{n}\right)=L_{\text{\ensuremath{\mathbf{x}}}}\left(I_{m},I_{n}\right)+\beta L_{\mathbf{q}}\left(I_{m},I_{n}\right),
\end{equation}
where $\beta$ is a hyperparameter to balance the rotational error and the translational error, since they tend to be in different scales.

\section{Experiments}
\label{sec:experiments}
Several experiments and their results are shown in this section. We begin this section by introducing the KITTI dataset, which is used to train our network. After that, we perform experiments on the Generator and the Discriminator to show how well the network is performing. Finally, we compare our results to ORB-SLAM2 \cite{murartal2017orb}, a well-known state-of-the-art feature-based method.

\subsection{Dataset}

The dataset used in this work is the well-known KITTI dataset \cite{geiger2013vision}. This dataset contains \num{22} sequences of images (commonly numbered from 00 to 21) acquired at 10 fps with a stereo camera located on the front of a car. They have been undistorted and rectified. The ground-truth poses for the sequences $\left\{ 00,01,\ldots10\right\}$ are provided in the form of a matrix $\mathbf{T}_{wk}\in\mathrm{SE}\left(3\right)$.

The images are preprocessed offline. Although the images were obtained with a stereo camera, we only take the images corresponding to the left camera, since we work in a monocular way. The resolution, initially from $1241\times376$ px is decreased to $128\times96$ px. To do this, the central region of the image is cropped with a size of $500\times375$ px, and then scaled to the final size. The main reason for this change is to reduce computation time. Additionally, by doing this we discard information that is irrelevant to the training, as is the case of the sky for example. After preprocessing the images, they are stored in pairs of consecutive frames along with the relative transformation between the two camera coordinate systems.
\subsubsection{Augmented dataset}
Having a large amount of data is one of the keys to train a CNN. In order to have more images, we generate new frames from the KITTI dataset. In the present work we mirror all the images for which we have ground-truth (the sequences $\left\{ 00,01,\ldots10\right\}$). Subsequently, these images are preprocessed to reduce their resolution to $128\times96$ px. Finally, the relative transformations can be obtained from the relative transformations of the original images. For each pair of mirror images $\left(I_{i}^{*},I_{i+1}^{*}\right)$ obtained from $\left(I_{i},I_{i+1}\right)$ with relative transformation 
\begin{equation}
\mathbf{T}_{i(i+1)}=\left[\begin{array}{cc}
\mathbf{R} & \mathbf{t}\\
\mathbf{0} & 1
\end{array}\right],
\end{equation}
the mirrored relative transformation $\mathbf{T}_{i(i+1)}^{*}$ is defined in equation \ref{eq:mirrored_transformation}:
\begin{equation} \label{eq:mirrored_transformation}
\mathbf{T}_{i(i+1)}^{*}=\left[\begin{array}{cc}
\mathbf{R^{*}} & \mathbf{t}\\
\mathbf{0} & 1
\end{array}\right],
\end{equation}
where $\mathbf{R^{*}}$ is defined in equation \ref{eq:mirrored_rotation}:
\begin{equation} \label{eq:mirrored_rotation}
\mathbf{R^{*}}=\mathbf{M}\mathbf{R\mathbf{M},} 
\end{equation}
and the matrix $\mathbf{M}$ is defined in equation \ref{eq:mirror_matrix}:
\begin{equation} \label{eq:mirror_matrix}
\mathbf{M}=\left[\begin{array}{ccc}
-1 & 0 & 0\\
0 & 1 & 0\\
0 & 0 & 1
\end{array}\right].
\end{equation}
\subsection{Generated images}
When working with generative models it is always interesting to observe the images generated by the Generator. Visually analyzing the output of this network is useful as a first indication that it is learning high level features on the observed images. The first experiment consists of training the network using the adversarial training between the Generator and Discriminator and without training the Discriminator for the pose regression. The sequences $\left\{00,01, \dots, 10 \right\}$ are chosen for the training. The size of the batch used is \num{100} pairs of frames and the network is trained during \num{20000} iterations. The Adam optimizer is used with a learning rate of $10^{-4}$, and in the cost function $L_{beta}$ the parameter $\beta$ is \num{100}.

\begin{figure}[!tbp]
    \centering
    \includegraphics[width=0.3\columnwidth]{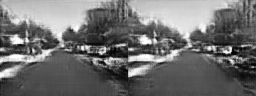}\\
    \vspace{0.15cm}
    \includegraphics[width=0.3\columnwidth]{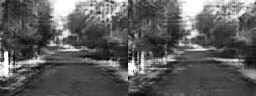}\\
    \vspace{0.15cm}
    \includegraphics[width=0.3\columnwidth]{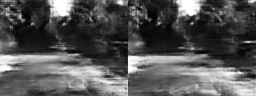}\\
    \caption{Pairs of frames generated by the proposed Generator. A slight motion that represents the camera motion can be observed in each pair.}
    \label{fig:samples_wgan_3}
\end{figure}

At the end of the training, it can be observed that the network generates consecutive image pairs that are very similar to the real ones. Figure~\ref{fig:samples_wgan_3} shows three pairs of images generated by the network. In each pair it is possible to visualize a slight motion between both images, as if it is a pair of consecutive frames obtained from the dataset. High level features are observed, such as the path in the center of the image and trees on either side of the path. In addition, the sky is easily distinguishable and the road goes towards the horizon.
\subsection{Semi-supervised training}
\label{subsec:sstrain}
This section analyses the influence of semi-supervised training on our work. Therefore, three different types of training are performed. The first one consists of performing the semi-supervised training proposed in Section \ref{sec:method}. This type of training implies to carry out the WGAN-GP characteristic adversarial training, and, at the end, to train the Discriminator to obtain the pose regression. To solve these tasks, the Discriminator receives as input two sources of information: images generated by the Generator in the first place, and then real images from the dataset. Secondly, it is proposed to train a CNN whose architecture is exactly the same as the Discriminator, except that in this case the probabilistic network output is removed. This network is trained to solve only the task of estimating the camera motion, ignoring the Generator and, therefore, without performing the adversarial training. Consequently, the network input only consists of real images. Finally, the third experiment consists of simultaneously performing, and in the same step, the adversarial training and the pose regression on the Discriminator. The configuration described in Section \ref{sec:method} is used in all experiments.

The KITTI error metric \cite{geiger2012are} is used to compare how much the estimated path differs from the actual path. The metric computes rotational and translational errors for several subsequences of lengths $\left\{100, 200, \dots, 800 \right\}$ meters. Each of these errors is averaged by the corresponding length and these are averaged by the number of subsequences for each of the lengths. In this way, the translational error and the rotational error are analyzed separately. The data used for the training are the sequences $\left\{00, 01, \dots,10\right\}$ and their mirrored sequences. All sequences are used for training except one sequence which is used for testing. In addition, the corresponding mirror sequence is removed from the training data. For example, the network is trained with the sequences $\left\{ 00, 01, \dots, 10 \right\}$ and their corresponding mirrored sequences and tested on the sequence 00. This is repeated for each of the sequences of the set $\left\{ 00, 01, \dots, 10 \right\}$. For the first case we train with \num{10000} iterations for the unsupervised part and \num{40000} iterations for the pose regression, while for the second and third cases we train with \num{50000} iterations. In each execution a batch of size \num{100} is used and the Adam optimizer is used with a learning rate of $10^{-4}$, and the cost function $L_{\beta}$ with parameter $\beta$=\num{100}.

Table \ref{tab:semi-sup-results} details the results obtained from this experiment according the KITTI metric. As can be seen, semi-supervised training in general improves the results obtained by the network. Sequence 01 represents a particular case in which errors are notably greater than in other sequences. This sequence is characterised by the fact that it is the only one with speeds over 60 km/h. Consequently, it is challenging for the network to estimate the trajectory accurately. In addition, the sequence takes place on a highway, and the components of the scene are far from the camera. The training data did not include movements at speeds similar to those in sequence 01, and it is likely that the performance of the network is affected by this. An interesting experiment would be adding pairs of frames to the training samples in which instead of using consecutive frames, pairs are formed by taking images each \emph{k} frames. Different values of \emph{k} could simulate different speeds and could eventually improve the results on sequence 01.

\begin{table}[!tbp]
\centering
\begin{adjustbox}{max width=\columnwidth}
\begin{tabular}{crrrrrr}
\toprule 
 & \multicolumn{2}{c}{WGANVO} & \multicolumn{2}{c}{Only-VO} & \multicolumn{2}{c}{Simult. GAN+VO}\\
 \cmidrule{2-7}
 & $t_{rel}$ & $r_{rel}$ & $t_{rel}$ & $r_{rel}$ & $t_{rel}$ & $r_{rel}$\\
Sequence & (\si{\percent}) & (\si[per-mode=symbol]{\degree\per100\metre}) 
& (\si{\percent}) & (\si[per-mode=symbol]{\degree\per100\metre})
& (\si{\percent}) & (\si[per-mode=symbol]{\degree\per100\metre})\\
\midrule
00 & \textbf{10.54} & 3.22 & 24.04 & 4.84 & 12.03 & \textbf{3.03}\\
01 & \textbf{24.94} & 3.54 & 26.34 & 4.07 & 45.29 & \textbf{3.21}\\
02 & 18.28 & \textbf{2.74} & \textbf{14.50} & 3.89 & 32.70 & 4.45\\
03 & 8.96 & 5.70 & \textbf{7.68} & \textbf{3.14} & 10.89 & 3.97\\
04 & \textbf{14.14} & 3.24 & 14.81 & 3.67 & 17.04 & \textbf{2.54}\\
05 & \textbf{7.01} & 3.85 & 9.18 & \textbf{2.51} & 9.22 & 2.64\\
06 & \textbf{7.87} & \textbf{2.19} & 11.26 & 2.97 & 30.16 & 6.68\\
07 & 7.71 & 3.79 & \textbf{6.52} & \textbf{3.74} & 18.10 & 4.78\\
08 & \textbf{9.04} & 3.85 & 9.94 & \textbf{3.59} & 15.14 & 3.81\\
09 & \textbf{10.49} & 4.91 & 13.65 & \textbf{2.85} & 36.46 & 4.83\\
10 & \textbf{10.89} & \textbf{5.03} & 12.91 & 5.54 & 17.44 & 6.64\\
\bottomrule

\end{tabular}
\end{adjustbox}
\caption{Comparison between different types of training. WGANVO stands for
the training proposed in Section \ref{sec:method}. Only-VO represents the training
that ignores adversarial training. Simult. GAN+VO represents the simultaneous (GAN and VO) minimization error. KITTI metric is used, and $t_{rel}$
and $r_{rel}$ represent translational and rotational error averaged
at intervals of 100 to 800 meters. The values representing the smallest error are shown in \textbf{bold}}
\label{tab:semi-sup-results}
\end{table}

\subsection{State-of-the-art methods}
ORB-SLAM2 \cite{murartal2017orb} is one of the most important feature-based SLAM method. For the present experiment the ORB-SLAM2 code is modified to deactivate the loop closure and the global optimization performed by the method. In that way, it is more similar to a VO algorithm. Then, the method is executed in its two versions, monocular and stereo, on the KITTI sequences $\left\{ 00, 01, \dots, 10 \right\}$ in their original resolution.

\begin{figure*}[!htbp]
  \centering
  \includegraphics[width=0.3\textwidth]{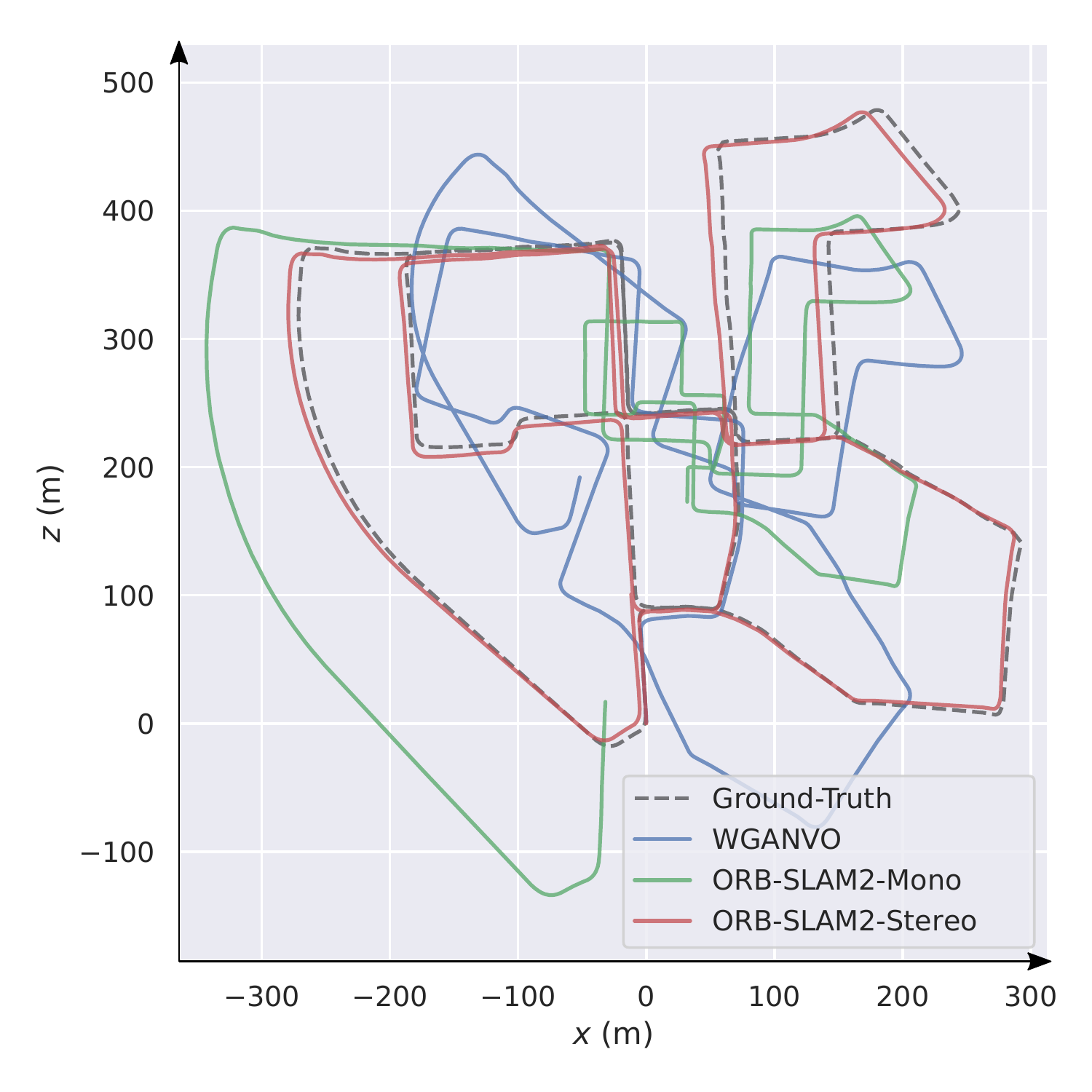}
  \includegraphics[width=0.3\textwidth]{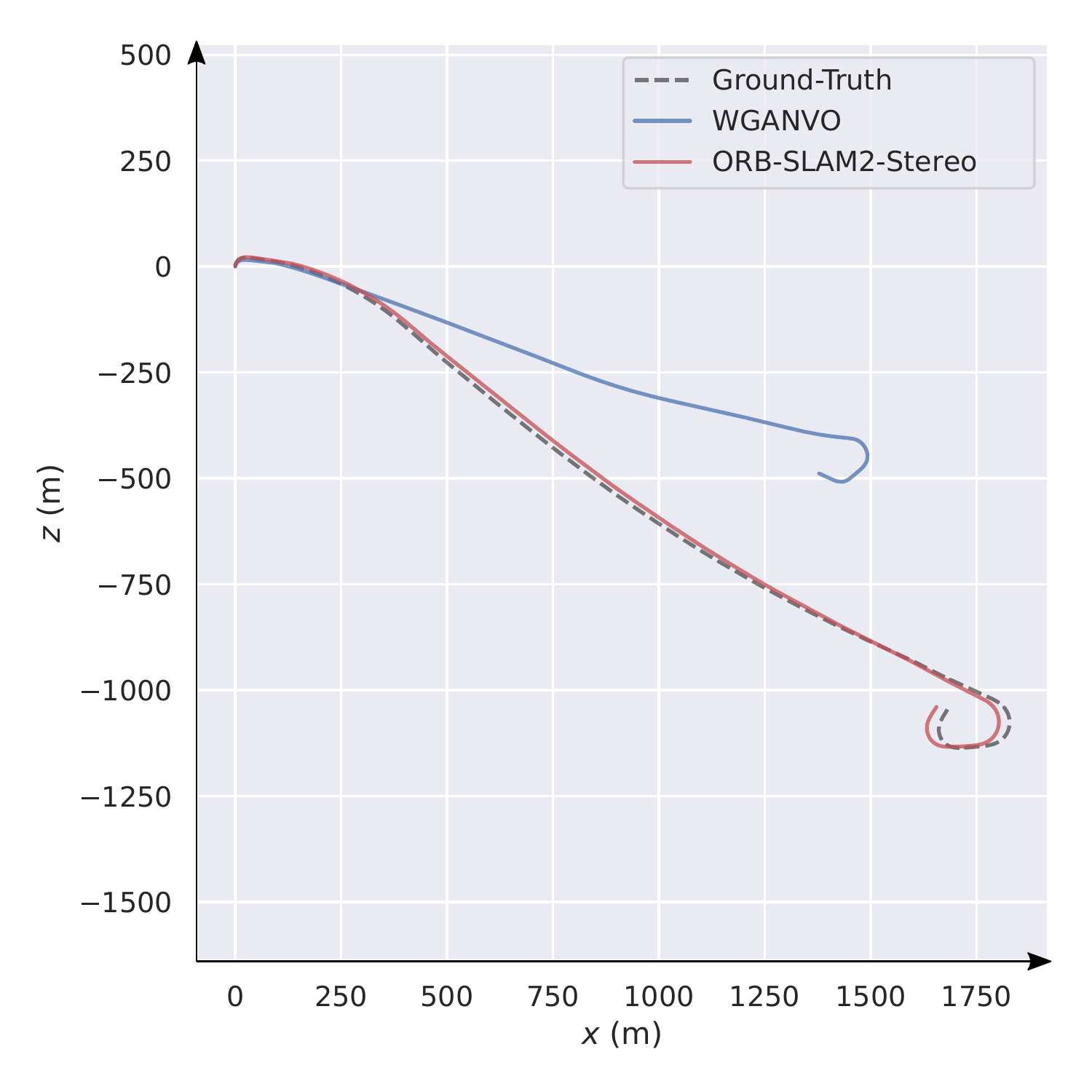}
  \includegraphics[width=0.3\textwidth]{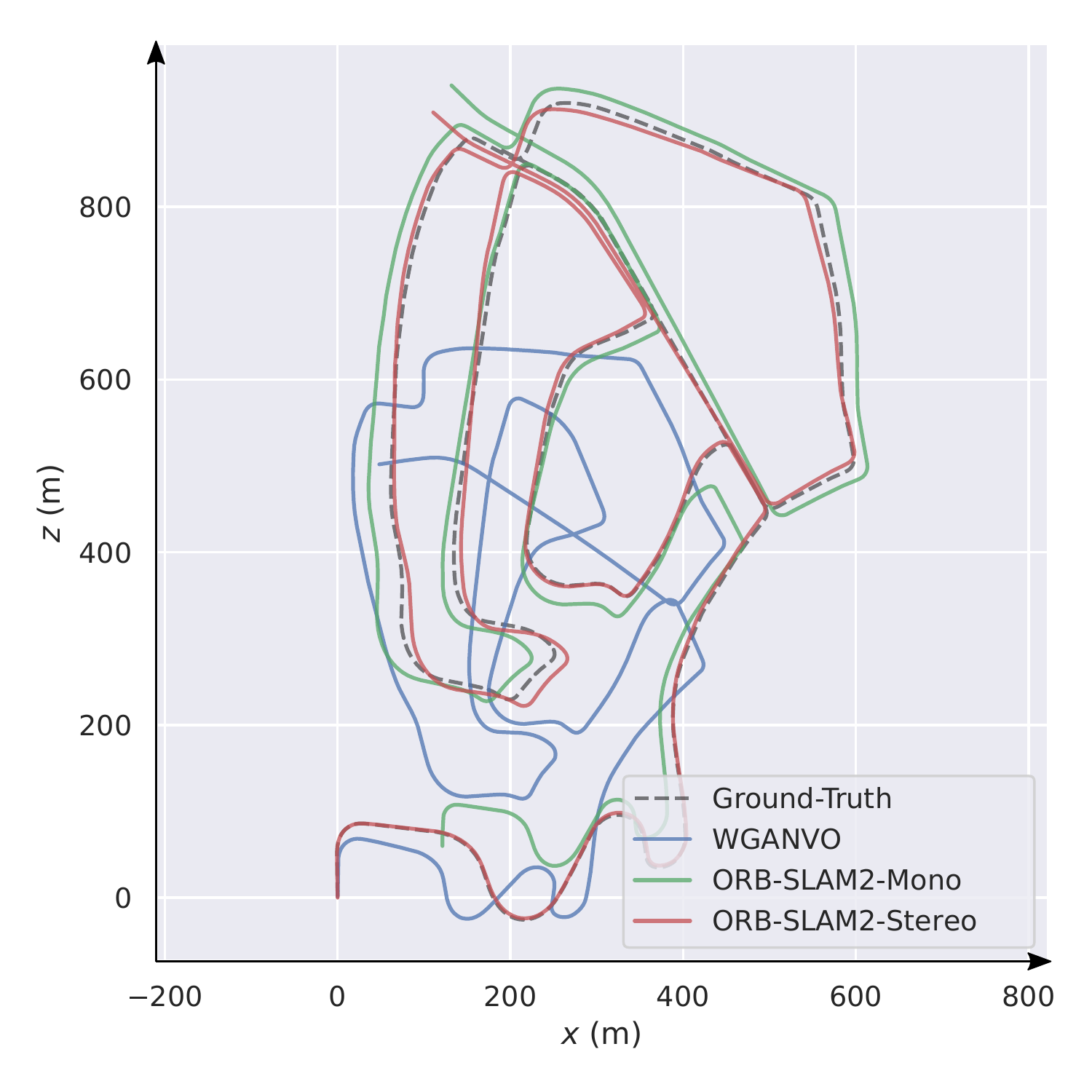}
  \\
  \includegraphics[width=0.3\textwidth]{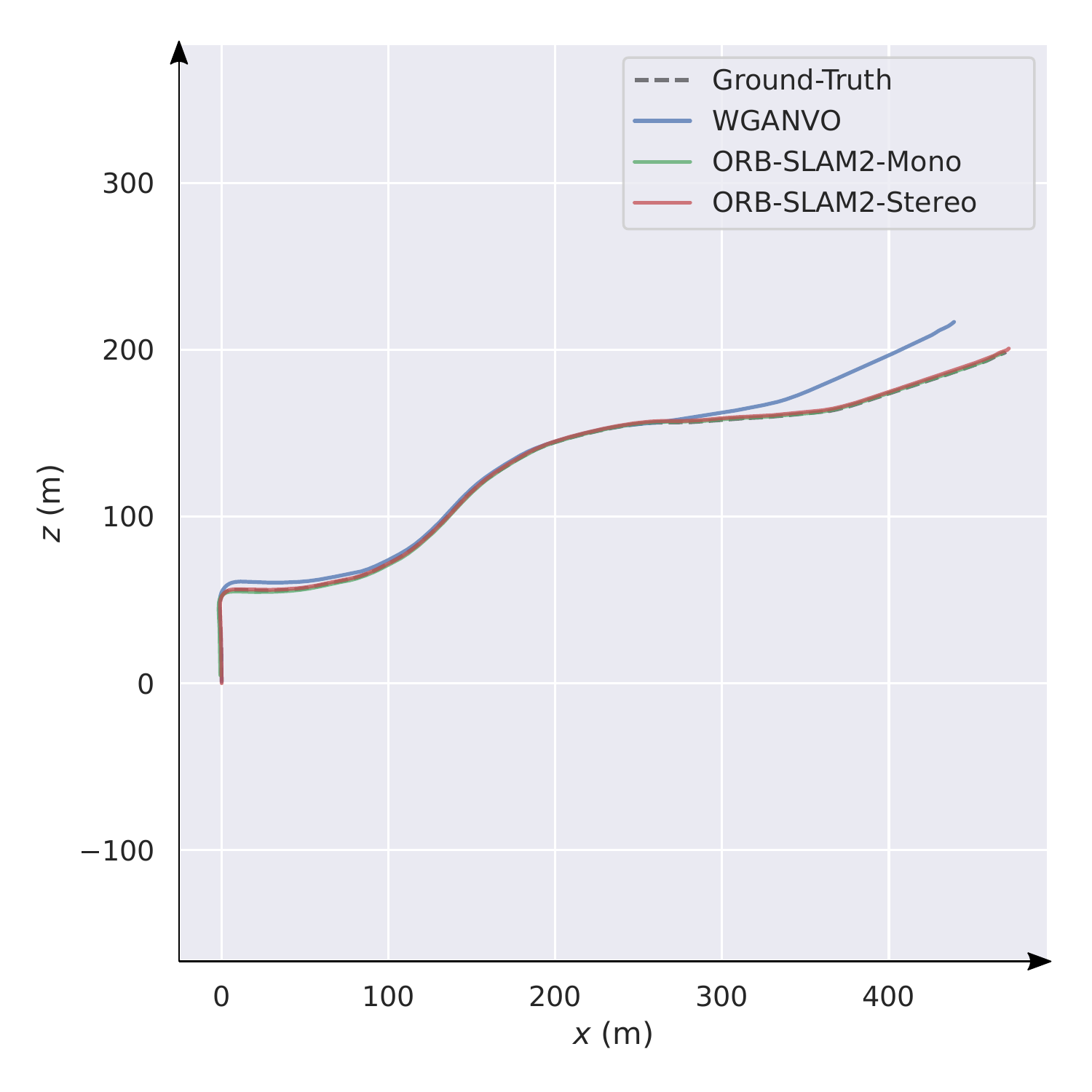}
  \includegraphics[width=0.3\textwidth]{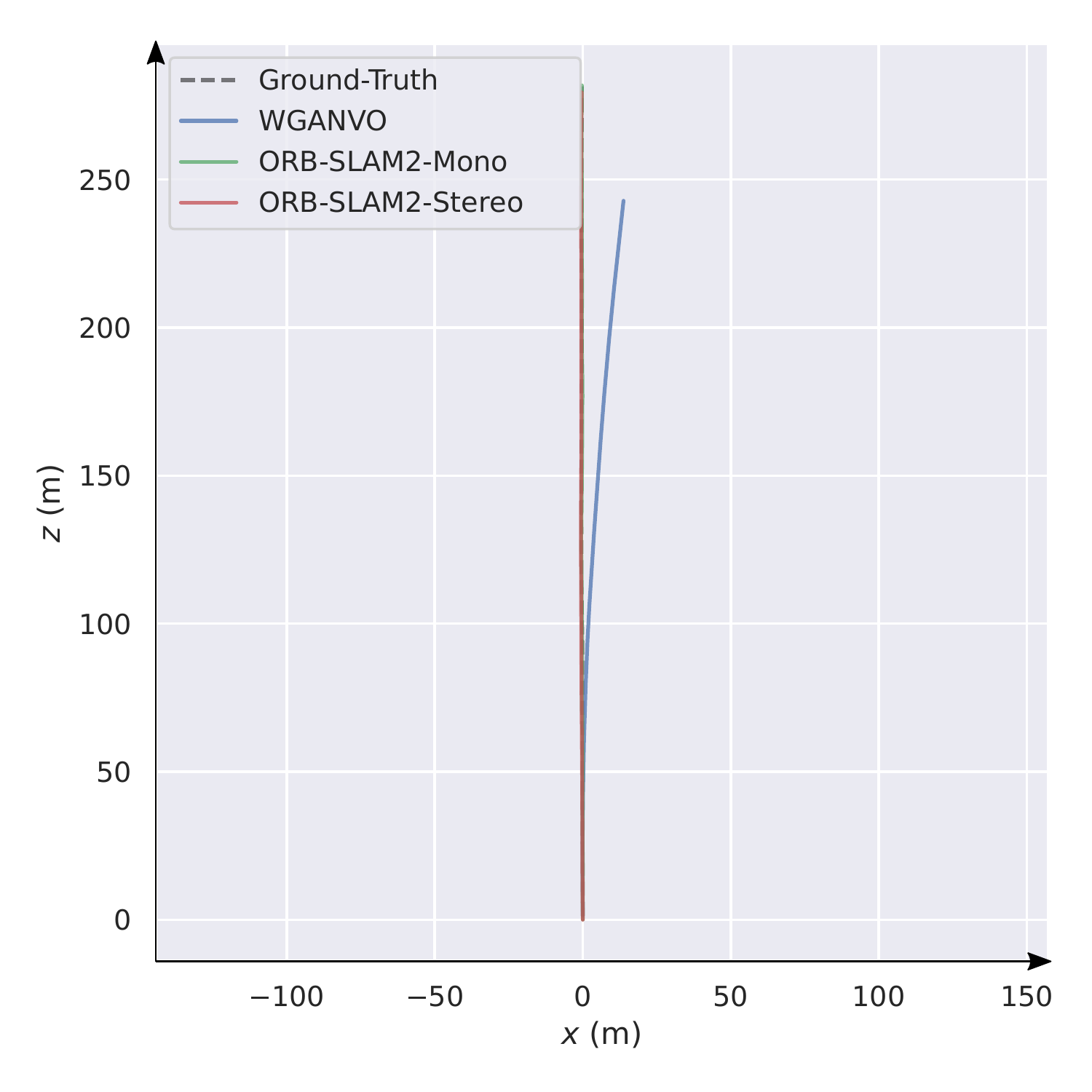}
  \includegraphics[width=0.3\textwidth]{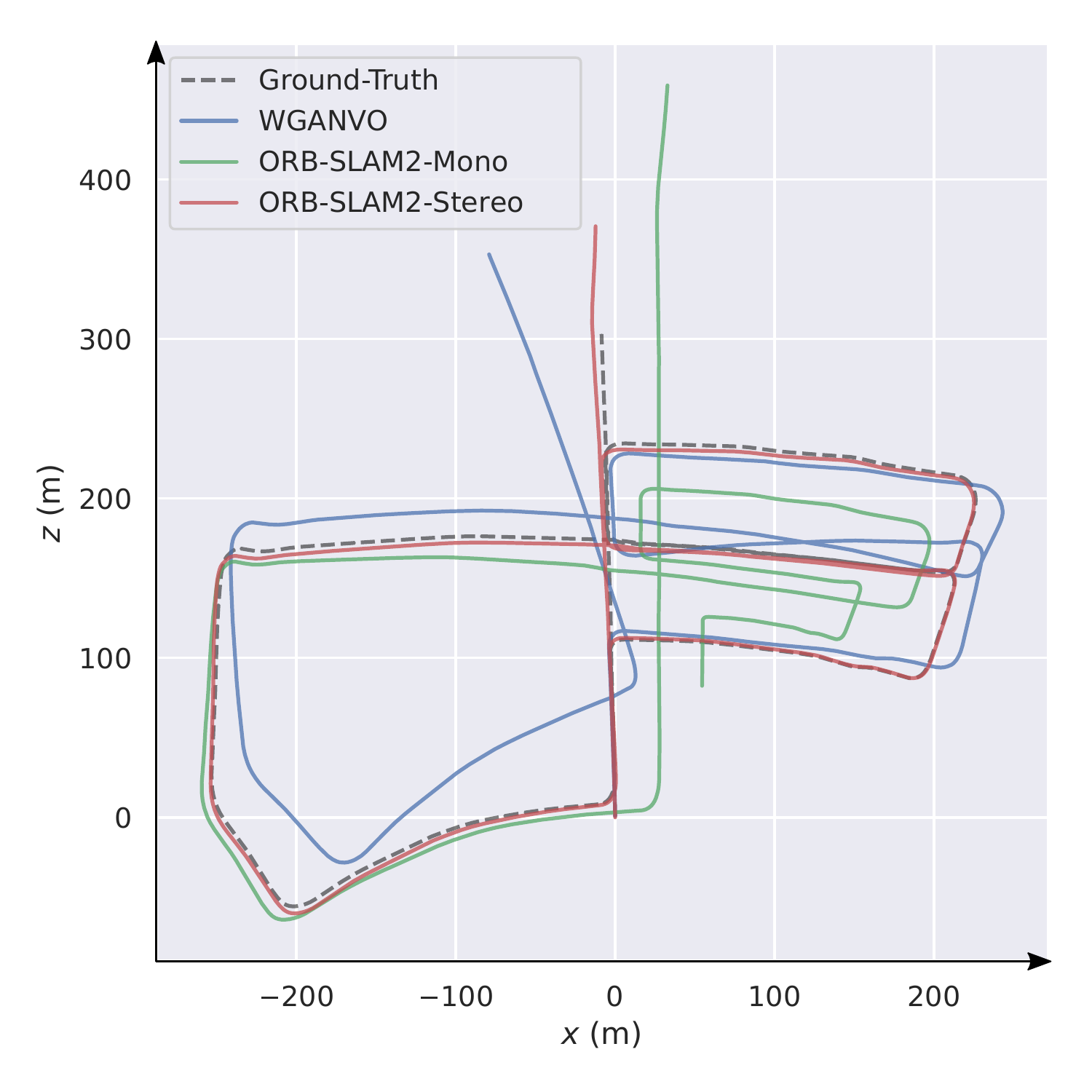}
  \\
  \includegraphics[width=0.3\textwidth]{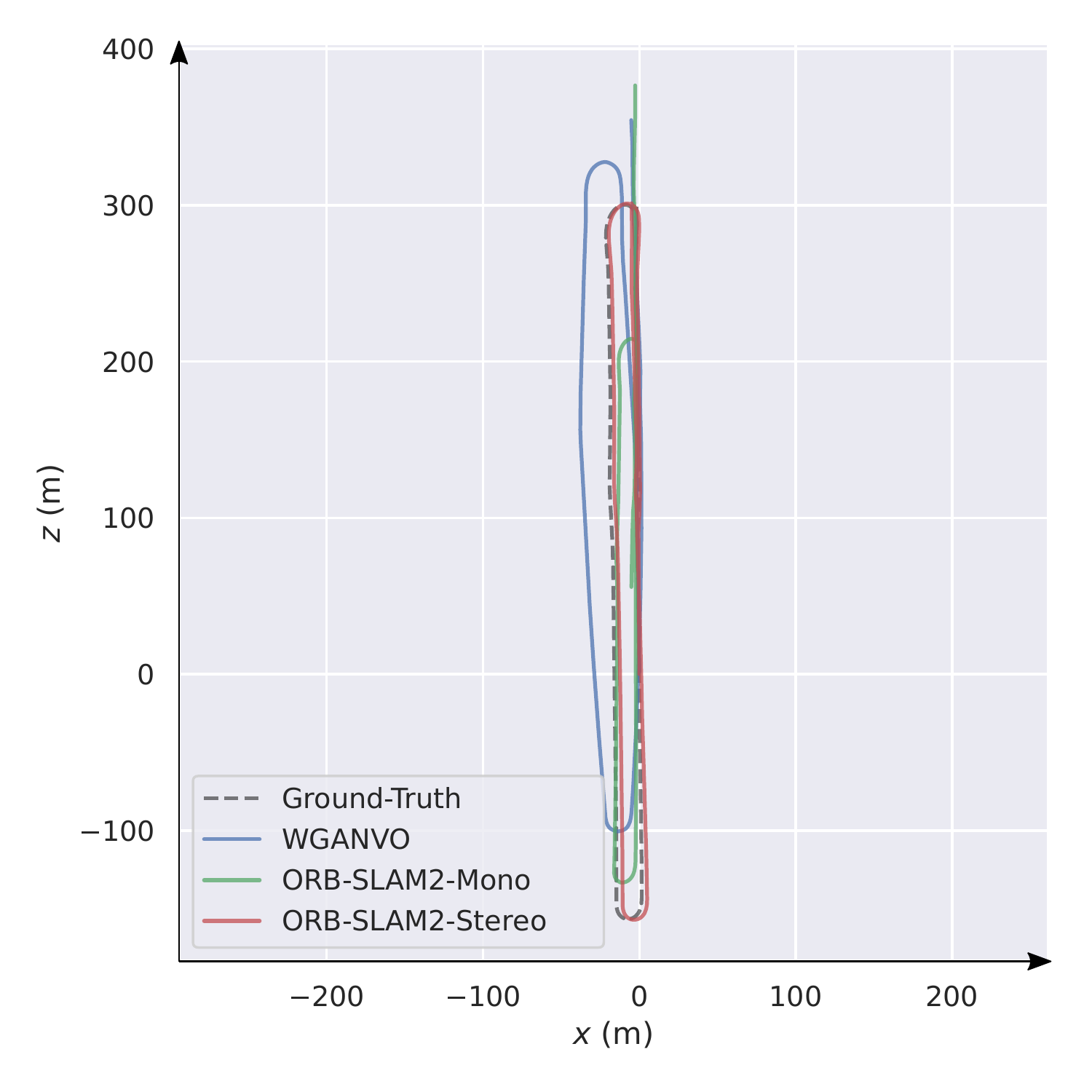}
  \includegraphics[width=0.3\textwidth]{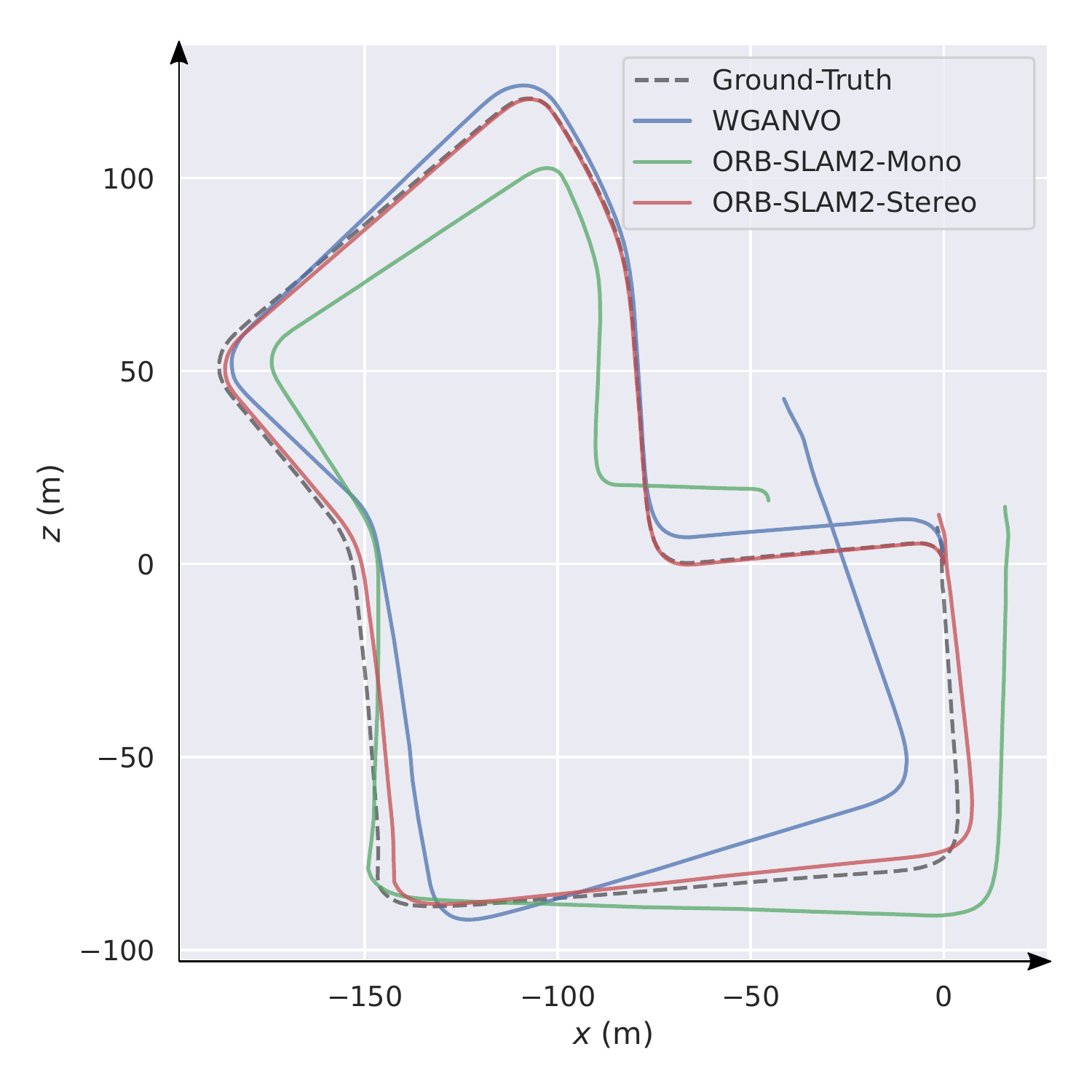}
  \includegraphics[width=0.3\textwidth]{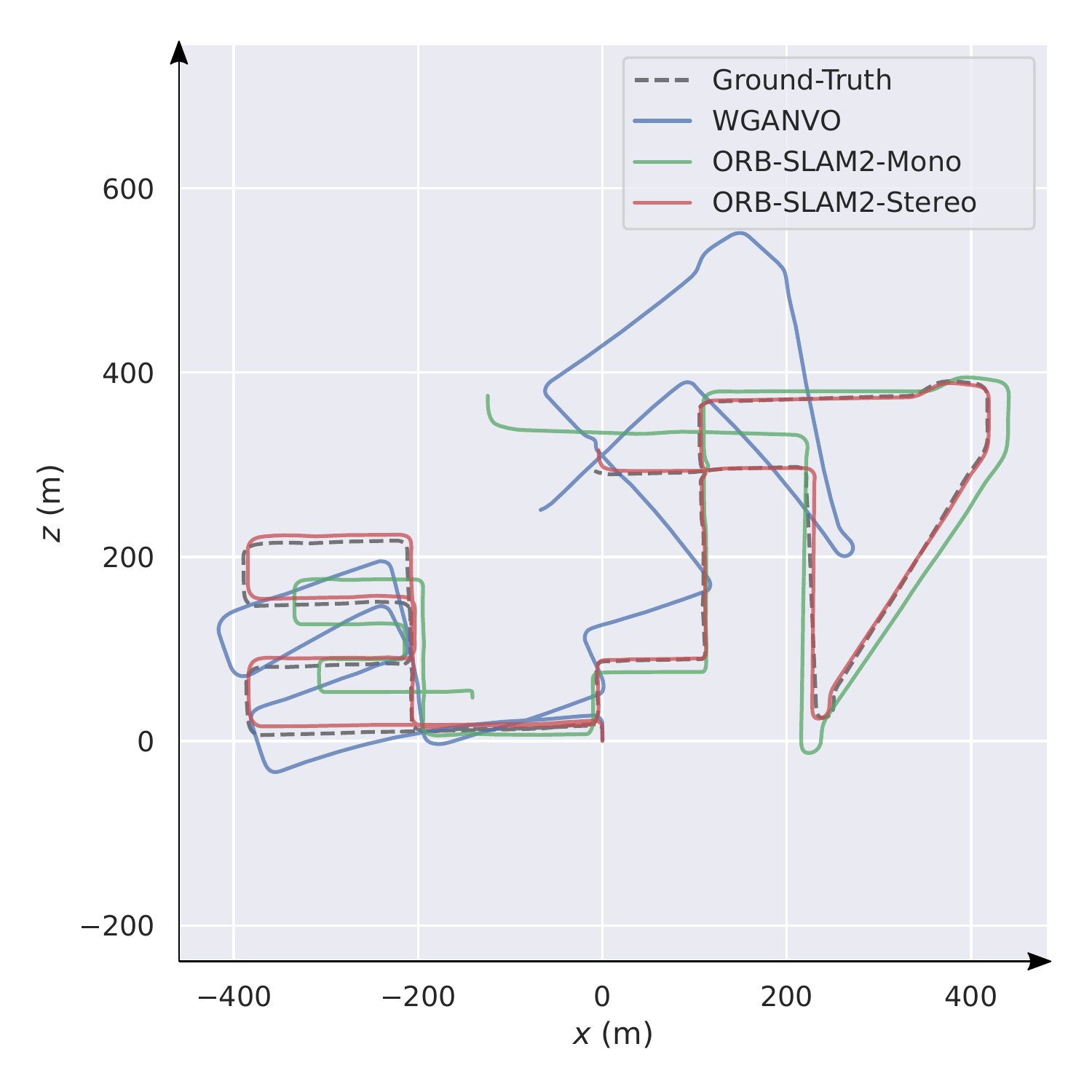}
  \\
  \includegraphics[width=0.3\textwidth]{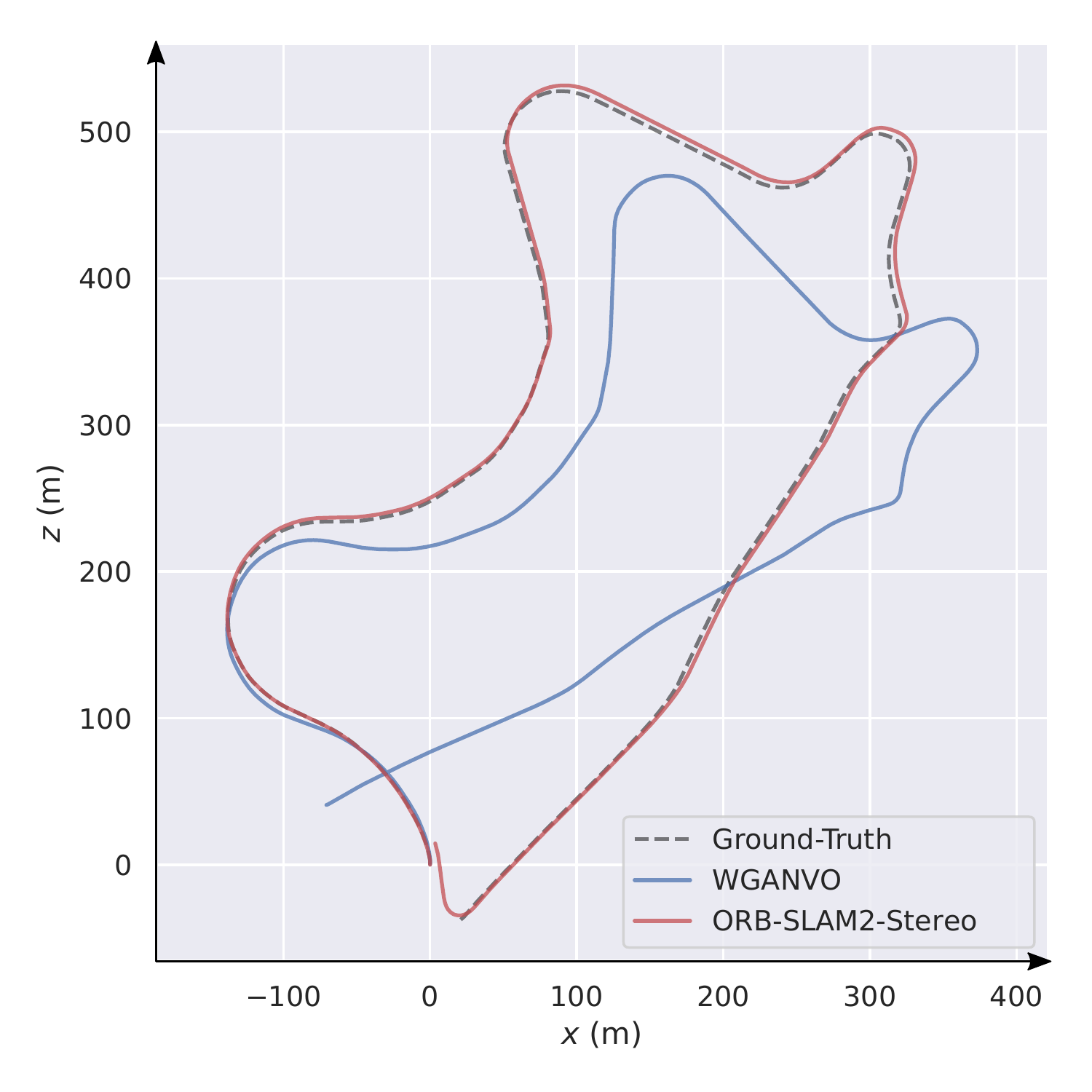}
  \includegraphics[width=0.3\textwidth]{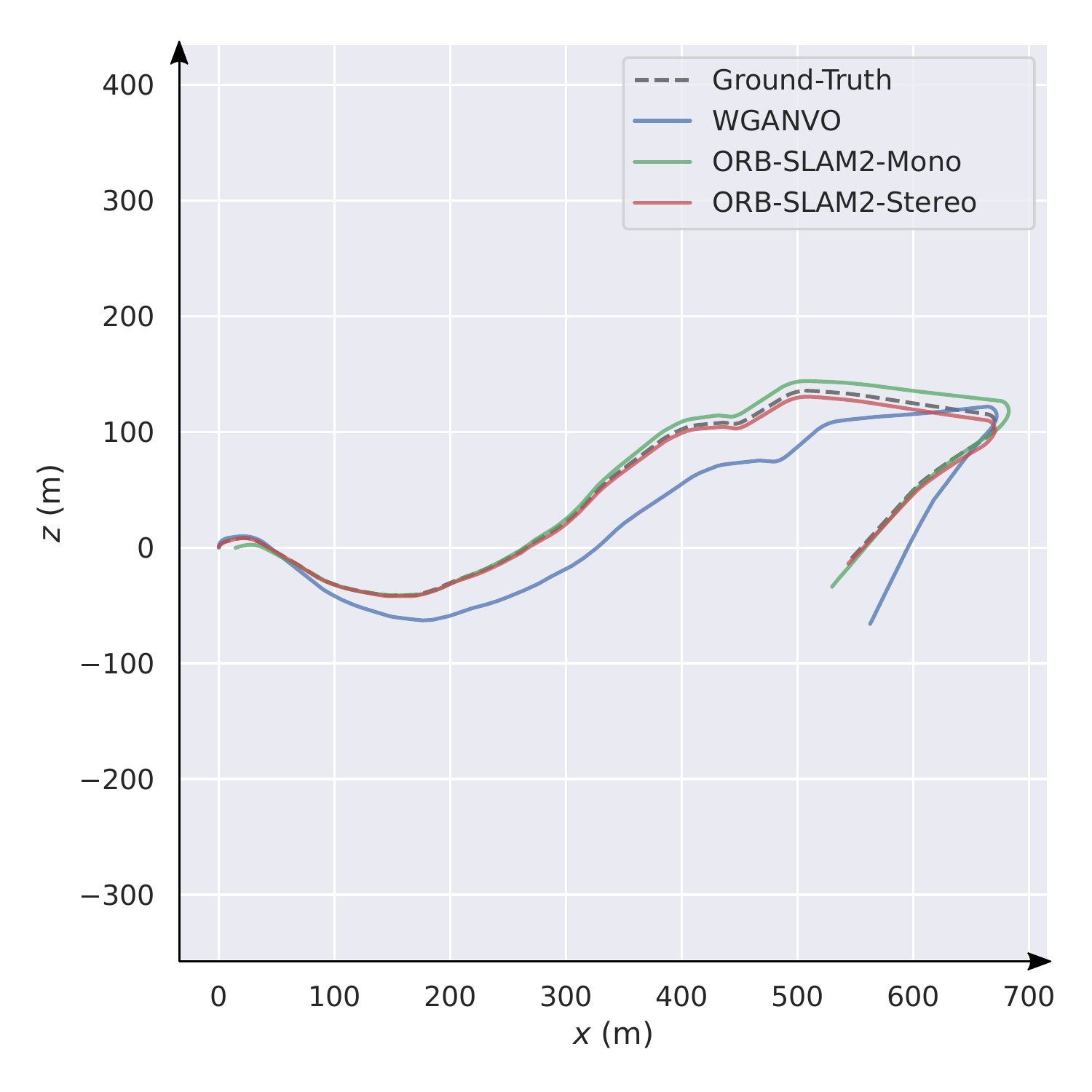}
  \caption{Comparison of the trajectory estimated by our method (blue) with ORB-SLAM2 Stereo (red), ORB-SLAM2 Mono (green) and the ground-truth for KITTI sequences 00 to 10, respectively.}
  \label{fig:wganvo-traj}
\end{figure*}

\begin{figure}[!tbp]
    \centering
    \includegraphics[width=0.6\columnwidth]{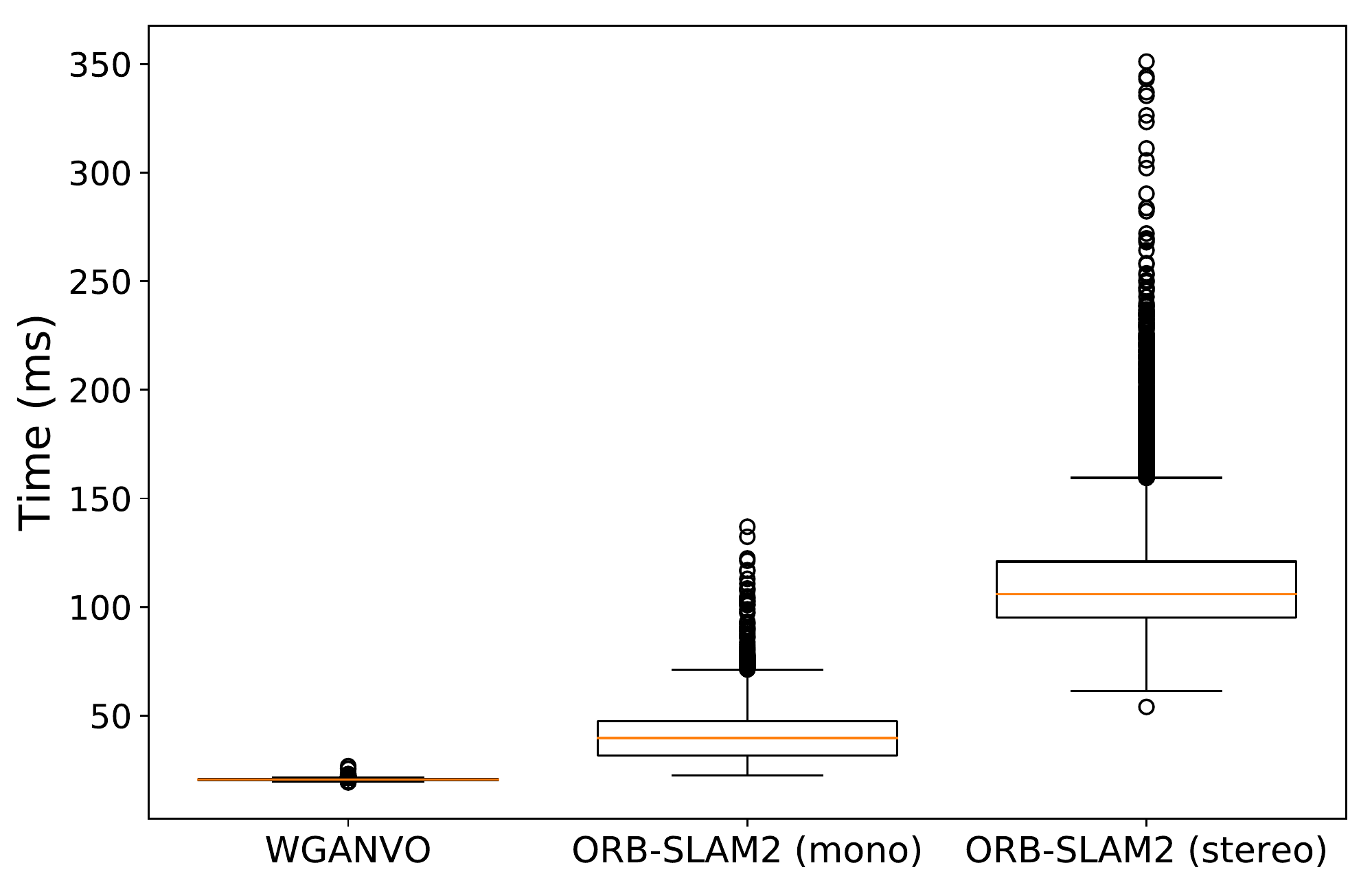}
    \caption{The following boxplots show the time taken to process each frame of KITTI sequences $\left\{00, 01, \dots,10\right\}$ in the executions shown in Fig. \ref{fig:wganvo-traj}.}
    \label{fig:times}
\end{figure}

Table \ref{tab:orb-slam} compares our method against the results obtained from ORB-SLAM2 runs. For the monocular case the trajectory was aligned and scaled with ground-truth using the Sim(3) Umeyama \cite{umeyama1991least} method.  It can be observed that although the method presents a great performance over KITTI in its stereo version, the accuracy decreases for the monocular version. Since monocular ORB-SLAM2 presents problems during the pose estimation for sequences 01 and 09, we do not report the error. As mentioned above, sequence 01 is often challenging as it has been recorded in a highway, at a faster speed and the points are far from the camera. Figure \ref{fig:wganvo-traj} shows the different trajectories obtained in this experiment. The three methods have been executed in a computer with Intel Core i7-3770 processor with 12 GB RAM and a GeForce GTX 770 GPU with 2 GB RAM. Additionally, Figure~\ref{fig:times} shows three boxplots plotted from the values (in milliseconds) the methods took to process each frame of the sequences $\left\{00, 01, \dots,10\right\}$.

To sum up, WGANVO has a lower performance than ORB-SLAM2 stereo. However, this is an expected result, as feature-based methods are the result of decades of research and this is reflected in their precision and robustness. On the other hand, Deep Learning techniques have only recently been applied in recent years to solve the VO problem. However, the accuracy of WGANVO is comparable to that of monocular ORB-SLAM2 for some of the sequences. Moreover, WGANVO does not suffer from the problem of not being able to recover the absolute scale of the scene, a general problem presented by the monocular methods based on features, since it can infer it from the training. In terms of execution times, WGANVO can process on average at a frame rate of 50 fps. KITTI dataset was recorded at 10 fps, so the method can process images obtained from a similar camera without any inconvenience.

\begin{table}[!tbp]
\centering
\begin{adjustbox}{max width=\columnwidth}
\begin{tabular}{crrrrrr}
\toprule 
 & \multicolumn{2}{c}{WGANVO} & \multicolumn{2}{c}{ORB-SLAM2} & \multicolumn{2}{c}{ORB-SLAM2}\\
  & \multicolumn{2}{c}{}  & \multicolumn{2}{c}{(Mono)} & \multicolumn{2}{c}{(Stereo)}\\
 \cmidrule{2-7}
 & $t_{rel}$ & $r_{rel}$ & $t_{rel}$ & $r_{rel}$ & $t_{rel}$ & $r_{rel}$\\
Sequence & (\si{\percent}) & (\si[per-mode=symbol]{\degree\per100\metre}) 
& (\si{\percent}) & (\si[per-mode=symbol]{\degree\per100\metre})
& (\si{\percent}) & (\si[per-mode=symbol]{\degree\per100\metre})\\
\midrule
00 & 10.54 & 3.22 & 23.01 & \textbf{0.30} & \textbf{0.88} & \textbf{0.30}\\
01 & 24.94 & 3.54 & - & - & \textbf{1.39} & \textbf{0.23}\\
02 & 18.28 & 2.74 & 6.63 & \textbf{0.24} & \textbf{0.81} & 0.28\\
03 & 8.96 & 5.70 & 1.12 & 0.19 & \textbf{0.73} & \textbf{0.17}\\
04 & 14.14 & 3.24 & 0.70 & 0.22 & \textbf{0.48} & \textbf{0.15}\\
05 & 7.01 & 3.85 & 12.34 & \textbf{0.22} & \textbf{0.61} & 0.25\\
06 & 7.87 & 2.19 & 17.71 & 0.27 & \textbf{0.76} & \textbf{0.23}\\
07 & 7.71 & 3.79 & 11.10 & \textbf{0.36} & \textbf{0.88} & 0.47\\
08 & 9.04 & 3.85 & 12.69 & \textbf{0.30} & \textbf{1.05} & 0.32\\
09 & 10.49 & 4.91 & - & - & \textbf{0.83} & \textbf{0.26}\\
10 & 10.89 & 5.03 & 3.90 & 0.30 & \textbf{0.57} & \textbf{0.26}\\
\bottomrule
\end{tabular}
\end{adjustbox}
\caption{Comparison between WGANVO and ORB-SLAM2 monocular and stereo, without loop closure or global optimization. KITTI metric is used, and $t_{rel}$ and $r_{rel}$ represent translational and rotational error averaged over intervals of 100 to 800 meters. Values representing the smallest error are shown in \textbf{bold}}
\label{tab:orb-slam}
\end{table}

\subsection{Alternative loss function}
An alternative loss function based on the reprojection error is defined in \cite{kendall2017geometric}. In particular, the term $L_{p}\left(I_{m},I_{n}\right)$ is defined in equation \ref{eq:repr_loss}:
\begin{equation} \label{eq:repr_loss}
L_{p}\left(I_{m},I_{n}\right)=\frac{1}{\left|G_{n}\right|}\sum_{\mathbf{x}\in G_{n}}\left\Vert f\left(\mathbf{R},\mathbf{x},\mathbf{t}\right)-f\left(\mathbf{\hat{R}},\mathbf{x},\mathbf{\hat{t}}\right)\right\Vert _{2},
\end{equation}
where the function $f\left(\mathbf{A},\mathbf{x},\mathbf{b}\right)$ is defined in equation \ref{eq:fun_f}:

\begin{equation} \label{eq:fun_f}
f\left(\mathbf{A},\mathbf{x},\mathbf{b}\right)=\pi\left(\mathbf{K}\left(\mathbf{A}\,\mathbf{x}+\mathbf{b}\right)\right),
\end{equation}

and $\pi$ is defined in equation \ref{eq:projection}:
\begin{equation} \label{eq:projection}
\pi\left(\left[a_{1}\ldots a_{n}\right]^{\top}\right)=\frac{1}{a_{n}}\left[\begin{array}{c}
a_{1}\\
\vdots\\
a_{n-1}
\end{array}\right],
\end{equation}
and $\mathbf{R}$ and $\mathbf{t}$ are the rotational and translational part, respectively, of the relative transformation between frame $I_{m}$ and frame $I_{n}$, $\mathbf{\hat{R}}$ and $\mathbf{\hat{t}}$ are the corresponding network estimates, $\mathbf{K}$ is the camera calibration matrix, and $G_{n}$ is a set of 3D points observable from image $I_{n}$ expressed in camera coordinates.

The 3D points have been obtained using KITTI stereo images. Features are extracted in the stereo pair using SIFT \cite{lowe1999object}. Subsequently, a matching is performed between the images of the pair in order to find points that correspond to the same 3D point of the observed scene. Finally, using calibration, a triangulation is performed using the DLT algorithm \cite{hartley2003multiple}.

Two different executions are carried out. The cost function $L_{\beta}$ with parameter $\beta=100$ is used on the one hand and $L_{p}$ on the other hand. Executions are performed by training in the same way as proposed in Section \ref{subsec:sstrain}, testing on each of the sequences of the set $\left\{ 00, 01, \dots, 10 \right\}$. In both cases the network is trained during \num{10000} iterations for the unsupervised part and \num{40000} iterations for the supervised part. In addition, a batch size of \num{100} is used and the Adam optimizer is used with a learning rate of $10^{-4}$.

Table \ref{tab:cost_func} shows the results obtained. As can be seen, no major differences are reported in favor of one or the other loss function. $L_{p}$ does not require the validation of a hyperparameter, while $L_{\beta}$ requires the validation of $\beta$. Additionally, in our work we found that the best results obtained with $L_{\beta}$ use a value $\beta=100$, and we apply it in all experiments, but according to the authors this value depends on the scene observed during the training \cite{kendall2017geometric}.

\begin{table}
\centering
\begin{tabular}[!tbp]{crrrr}
\toprule 
 & \multicolumn{2}{c}{$L_{\beta}$} & \multicolumn{2}{c}{$L_{p}$}\\
\cmidrule{2-5}
 & $t_{rel}$ & $r_{rel}$ & $t_{rel}$ & $r_{rel}$\\
Sequence & (\si{\percent}) & (\si[per-mode=symbol]{\degree\per100\metre}) & (\si{\percent}) & (\si[per-mode=symbol]{\degree\per100\metre})\\
\midrule
00 & \textbf{10.54} & \textbf{3.22} & 11.01 & 4.41\\
01 & 24.94 & 3.54 & \textbf{23.31} & \textbf{3.27}\\
02 & 18.28 & \textbf{2.74} & \textbf{16.11} & 2.88\\
03 & \textbf{8.96} & 5.70 & 9.68 & \textbf{3.41}\\
04 & \textbf{14.14} & \textbf{3.24} & 14.91 & 3.71\\
05 & \textbf{7.01} & \textbf{3.85} & 7.18 & 3.91\\
06 & 7.87 & \textbf{2.19} & \textbf{7.56} & 2.37\\
07 & \textbf{7.71} & 3.79 & 7.82 & \textbf{3.74}\\
08 & \textbf{9.04} & 3.85 & 9.41 & \textbf{3.19}\\
09 & \textbf{10.49} & 4.91 & 10.5 & \textbf{3.85}\\
10 & \textbf{10.89} & \textbf{5.03} & 10.97 & 5.41\\
\bottomrule
\end{tabular}
\caption{Comparison between the two loss functions $L_{\beta}$ and $L_{p}$. The KITTI metric is used, and $t_{rel}$ and $r_{rel}$ represent translational and rotational error averaged over 100 to 800 meter intervals. Values representing the smallest error are shown in \textbf{bold}\label{tab:cost_func}}
\end{table}

\section{Conclusions}
\label{sec:conclusions}
This paper presents WGANVO, a Visual Odometry method based on CNN that allows estimating the camera motion. A particular type of CNN is used whose architecture is based on GAN. More specifically, a variant of GAN known as WGAN-GP \cite{gulrajani2017improved} is implemented, as it is more stable during training than the original GAN formulation. The semi-supervised training of our method consists in performing the adversarial training minimizing the original cost function of WGAN-GP and then minimizing the cost function of VO in a supervised manner using the ground-truth. Consequently, the Discriminator learns to distinguish high level visual characteristics from labeled and unlabeled data. It is worth mentioning that our method does not present the problem of not being able to recover the absolute scale of the scene, a general problem presented by geometry based monocular systems, since it infers it during the supervised part of the training.

As future work, it is proposed to add the camera calibration parameters during the training stage in order to generate an model independent of the camera used to capture the training images. In that way it is possible to train and test the algorithm on different datasets, and even train with datasets with different cameras without affecting performance. CAM-Convs \cite{facil2019camconvs} set a precedent in this aspect, since they estimate the depth of a scene generating an model independent of the camera, by concatenating the calibration parameters with the convolutional features maps. 

\section*{Acknowledgements}
This work is part of the \textit{Development of a weed remotion mobile robot} project at CIFASIS (CONICET-UNR).

\bibliographystyle{ieeetr}
\bibliography{biblio}

\begin{thebibliography}{10}

\bibitem{thrun2005probabilistics}
S.~Thrun, W.~Burgard, and D.~Fox, {\em {Probabilistic Robotics}}.
\newblock (The MIT Press. 2005).

\bibitem{siciliano2016handbook}
B.~Siciliano and O.~Khatib, {\em Springer Handbook of Robotics}.
\newblock (Springer Publishing Company, Incorporated. 2016).

\bibitem{nister2004visual}
D.~Nist{\'e}r, O.~Naroditsky, and J.~Bergen,
  ``\href{http://dx.doi.org/10.1109/CVPR.2004.1315094}{Visual odometry},'' in
  {\em Proceedings of the IEEE Conference on Computer Vision and Pattern
  Recognition (2004)}, pp.~652--659.

\bibitem{scaramuzza2011visual}
D.~{Scaramuzza} and F.~{Fraundorfer},
  ``\href{http://dx.doi.org/10.1109/MRA.2011.943233}{Visual Odometry
  [Tutorial]},'' {\em {IEEE} Robotics and Automation Magazine (2011)},
  pp.~80--92.

\bibitem{comport2010realtime}
A.~I. Comport, E.~Malis, and P.~Rives,
  ``\href{https://doi.org/10.1177/0278364909356601}{Real-time quadrifocal
  visual odometry},'' {\em International Journal of Robotics Research (2010)},
  pp.~245--266.

\bibitem{krombach2016combining}
N.~Krombach, D.~Droeschel, and S.~Behnke,
  ``\href{https://dx.doi.org/10.1007/978-3-319-48036-7_62}{Combining
  Feature-based and Direct Methods for Semi-dense Real-time Stereo Visual
  Odometry},'' in {\em Proceedings of the International Conference on
  Intelligent Autonomous Systems (2016)}, pp.~855--868.

\bibitem{engel2018dso}
J.~Engel, V.~Koltun, and D.~Cremers,
  ``\href{http://dx.doi.org/10.1109/TPAMI.2017.2658577}{Direct Sparse
  Odometry},'' {\em {IEEE} Transactions on Pattern Analysis and Machine
  Intelligence (2018)}, pp.~611--625.

\bibitem{forster2014svo}
C.~Forster, M.~Pizzoli, and D.~Scaramuzza,
  ``\href{http://dx.doi.org/10.1109/ICRA.2014.6906584}{SVO: Fast semi-direct
  monocular visual odometry},'' in {\em Proceedings of the IEEE International
  Conference on Robotics and Automation (2014)}, pp.~15--22.

\bibitem{pire2017sptam}
T.~Pire, T.~Fischer, G.~Castro, P.~De~Crist{\'o}foris, J.~Civera, and
  J.~Jacobo~Berlles,
  ``\href{http://dx.doi.org/10.1016/j.robot.2017.03.019}{S-PTAM: Stereo
  Parallel Tracking and Mapping},'' {\em Journal of Robotics and Autonomous
  Systems (2017)}, pp.~27--42.

\bibitem{lecun2015deep}
Y.~LeCun, Y.~Bengio, and G.~Hinton,
  ``\href{https://www.nature.com/articles/nature14539}{Deep learning},'' {\em
  Nature}, vol.~521, no.~7553, p.~436, (2015).

\bibitem{salimans2016improved}
T.~Salimans, I.~Goodfellow, W.~Zaremba, V.~Cheung, A.~Radford, and X.~Chen,
  ``\href{http://dl.acm.org/citation.cfm?id=3157096.3157346}{Improved
  Techniques for Training GANs},'' in {\em Proceedings of the International
  Conference on Neural Information Processing Systems (2016)}, pp.~2234--2242.

\bibitem{krizhevsky12alexnet}
A.~Krizhevsky, I.~Sutskever, and G.~E. Hinton,
  ``\href{http://papers.nips.cc/paper/4824-imagenet-classification-with-deep-convolutional-neural-networks.pdf}{ImageNet
  Classification with Deep Convolutional Neural Networks},'' in {\em
  Proceedings of the Advances in Neural Information Processing Systems (2012)},
  pp.~1097--1105.

\bibitem{goodfellow2014generative}
I.~Goodfellow, J.~Pouget-Abadie, M.~Mirza, B.~Xu, D.~Warde-Farley, S.~Ozair,
  A.~Courville, and Y.~Bengio,
  ``\href{http://papers.nips.cc/paper/5423-generative-adversarial-nets.pdf}{Generative
  adversarial nets},'' in {\em Proceedings of the Advances in Neural
  Information Processing Systems (2014)}, pp.~2672--2680.

\bibitem{radford2015unsupervised}
A.~Radford, L.~Metz, and S.~Chintala, ``{Unsupervised Representation Learning
  with Deep Convolutional Generative Adversarial Networks},'' in {\em
  Proceedings of the International Conference on Learning Representations
  (2016)}.

\bibitem{wang2017deepvo}
S.~Wang, R.~Clark, H.~Wen, and N.~Trigoni,
  ``\href{http://dx.doi.org/10.1109/ICRA.2017.7989236}{DeepVO: Towards
  end-to-end visual odometry with deep Recurrent Convolutional Neural
  Networks},'' in {\em Proceedings of the IEEE International Conference on
  Robotics and Automation (2017)}, pp.~2043--2050.

\bibitem{tateno2017cnn}
K.~Tateno, F.~Tombari, I.~Laina, and N.~Navab,
  ``\href{http://dx.doi.org/10.1109/CVPR.2017.695}{CNN-SLAM: Real-Time Dense
  Monocular SLAM with Learned Depth Prediction},'' in {\em Proceedings of the
  IEEE Conference on Computer Vision and Pattern Recognition (2017)},
  pp.~6565--6574.

\bibitem{agrawal2015learning}
P.~Agrawal, J.~Carreira, and J.~Malik,
  ``\href{http://dx.doi.org/10.1109/ICCV.2015.13}{Learning to See by Moving},''
  in {\em Proceedings of the International Conference on Computer Vision
  (2015)}, pp.~37--45.

\bibitem{murartal2015orb}
R.~Mur-Artal, J.~M.~M. Montiel, and J.~D. Tard\'{o}s,
  ``\href{http://dx.doi.org/10.1109/TRO.2015.2463671}{ORB-SLAM: A Versatile and
  Accurate Monocular SLAM System},'' {\em {IEEE} Transactions on Robotics
  (2015)}, pp.~1147--1163.

\bibitem{gulrajani2017improved}
I.~Gulrajani, F.~Ahmed, M.~Arjovsky, V.~Dumoulin, and A.~C. Courville,
  ``\href{http://papers.nips.cc/paper/7159-improved-training-of-wasserstein-gans.pdf}{Improved
  Training of Wasserstein GANs},'' in {\em Proceedings of the Advances in
  Neural Information Processing Systems (2017)}, pp.~5767--5777.

\bibitem{geiger2013vision}
A.~Geiger, P.~Lenz, C.~Stiller, and R.~Urtasun,
  ``\href{http://dx.doi.org/10.1177/0278364913491297}{Vision Meets Robotics:
  The KITTI Dataset},'' {\em International Journal of Robotics Research
  (2013)}, pp.~1231--1237.

\bibitem{murartal2017orb}
R.~{Mur-Artal} and J.~D. Tard\'{o}s,
  ``\href{http://dx.doi.org/10.1109/TRO.2017.2705103}{ORB-SLAM2: An Open-Source
  SLAM System for Monocular, Stereo, and RGB-D Cameras},'' {\em {IEEE}
  Transactions on Robotics (2017)}, pp.~1255--1262.

\bibitem{engel2014lsd}
J.~Engel, T.~Sch{\"o}ps, and D.~Cremers, ``{LSD-SLAM: Large-Scale Direct
  Monocular SLAM},'' in {\em Proceedings of the European Conference on Computer
  Vision (2014)}, pp.~834--849.

\bibitem{yang2018deep}
N.~Yang, R.~Wang, J.~St{\"u}ckler, and D.~Cremers,
  ``\href{http://dx.doi.org/10.1007/978-3-030-01237-3_50}{Deep Virtual Stereo
  Odometry: Leveraging Deep Depth Prediction for Monocular Direct Sparse
  Odometry},'' in {\em Proceedings of the European Conference on Computer
  Vision (2018)}, pp.~835--852.

\bibitem{li2017undeepvo}
R.~Li, S.~Wang, Z.~Long, and D.~Gu,
  ``\href{http://dx.doi.org/10.1109/ICRA.2018.8461251}{UnDeepVO: Monocular
  Visual Odometry through Unsupervised Deep Learning},'' in {\em Proceedings of
  the IEEE International Conference on Robotics and Automation (2018)},
  pp.~7286--7291.

\bibitem{almalioglu2018ganvo}
Y.~Almalioglu, M.~R.~U. Saputra, P.~P.~B. de~Gusmao, A.~Markham, and
  N.~Trigoni, ``\href{https://dx.doi.org/10.1109/ICRA.2019.8793512}{GANVO:
  Unsupervised Deep Monocular Visual Odometry and Depth Estimation with
  Generative Adversarial Networks},'' in {\em Proceedings of the IEEE
  International Conference on Robotics and Automation (2019)}, pp.~5474--5480.

\bibitem{yin2018geonet}
Z.~Yin and J.~Shi, ``\href{http://dx.doi.org/10.1109/CVPR.2018.00212}{GeoNet:
  Unsupervised Learning of Dense Depth, Optical Flow and Camera Pose},'' in
  {\em Proceedings of the IEEE Conference on Computer Vision and Pattern
  Recognition (2018)}, pp.~1983--1992.

\bibitem{kendall2015posenet}
A.~Kendall, M.~Grimes, and R.~Cipolla,
  ``\href{http://dx.doi.org/10.1109/ICCV.2015.336}{PoseNet: A Convolutional
  Network for Real-Time 6-DOF Camera Relocalization},'' in {\em Proceedings of
  the International Conference on Computer Vision (2015)}, pp.~2938--2946.

\bibitem{szegedy2015going}
C.~Szegedy, W.~Liu, Y.~Jia, P.~Sermanet, S.~Reed, D.~Anguelov, D.~Erhan,
  V.~Vanhoucke, and A.~Rabinovich,
  ``\href{http://dx.doi.org/10.1109/CVPR.2015.7298594}{Going Deeper with
  Convolutions},'' in {\em Proceedings of the IEEE Conference on Computer
  Vision and Pattern Recognition (2015)}, pp.~1--9.

\bibitem{kendall2017geometric}
A.~Kendall and R.~Cipolla,
  ``\href{http://dx.doi.org/10.1109/CVPR.2017.694}{Geometric Loss Functions for
  Camera Pose Regression with Deep Learning},'' in {\em Proceedings of the IEEE
  Conference on Computer Vision and Pattern Recognition (2017)},
  pp.~6555--6564.

\bibitem{geiger2012are}
A.~Geiger, P.~Lenz, and R.~Urtasun,
  ``\href{http://dx.doi.org/10.1109/CVPR.2012.6248074}{Are we ready for
  Autonomous Driving? The KITTI Vision Benchmark Suite},'' in {\em Proceedings
  of the IEEE Conference on Computer Vision and Pattern Recognition (2012)},
  pp.~3354--3361.

\bibitem{umeyama1991least}
S.~{Umeyama}, ``\href{http://dx.doi.org/10.1109/34.88573}{Least-squares
  estimation of transformation parameters between two point patterns},'' {\em
  {IEEE} Transactions on Pattern Analysis and Machine Intelligence (1991)},
  pp.~376--380.

\bibitem{lowe1999object}
D.~G. Lowe, ``\href{http://dx.doi.org/10.1109/ICCV.1999.790410}{Object
  recognition from local scale-invariant features},'' in {\em Proceedings of
  the International Conference on Computer Vision (1999)}, pp.~1150--1157.

\bibitem{hartley2003multiple}
R.~Hartley and A.~Zisserman, {\em {Multiple View Geometry in Computer Vision}}.
\newblock Cambridge University Press, New York, USA, (2003).

\bibitem{facil2019camconvs}
J.~M. Facil, B.~Ummenhofer, H.~Zhou, L.~Montesano, T.~Brox, and J.~Civera,
  ``\href{http://dx.doi.org/10.1109/CVPR.2019.01210}{CAM-Convs: Camera-Aware
  Multi-Scale Convolutions for Single-View Depth},'' in {\em Proceedings of the
  IEEE Conference on Computer Vision and Pattern Recognition (2019)},
  pp.~11818--11827.

\end{thebibliography}

\end{document}